%% file: neurips_2026.tex
\documentclass{article}

\usepackage[sglblindworkshop, final]{neurips_2026}
\usepackage{wrapfig}
\usepackage{subcaption}
\workshoptitle{ML for Systems}

\usepackage[utf8]{inputenc} 
\usepackage[T1]{fontenc}    
\usepackage{hyperref}       
\usepackage{url}            
\usepackage{booktabs}       
\usepackage{amsfonts}       
\usepackage{nicefrac}       
\usepackage{microtype}      
\usepackage{listings}
\usepackage{xcolor}         

\usepackage[ruled,vlined]{algorithm2e}
\usepackage{mathtools}
\usepackage{bm}
\usepackage{esvect}
\usepackage{array}
\usepackage{enumitem}
\usepackage{subcaption}
\usepackage{caption}
\usepackage{soul}
\usepackage{amssymb}
\usepackage{graphicx}

\usepackage{subcaption}
\usepackage{amsmath}
\newcommand{\name}{{\tt Rift}}
\newcommand{\para}[1]{\vspace{6pt}\noindent{\bf#1}}

\newcommand{\textred}[1]{\textcolor{red}{#1}}
\ifx\noeditingmarks\undefined
   \newcommand{\pgwrapper}[2]{\textred{#1: #2}}
\else
   \newcommand{\pgwrapper}[2]{}
\fi

\newcommand{\squishlist}
{
    \begin{list}{$\bullet$}
    {
        \setlength{\itemsep}{0pt}      \setlength{\parsep}{3pt}
        \setlength{\topsep}{3pt}       \setlength{\partopsep}{0pt}
        \setlength{\leftmargin}{1.5em} \setlength{\labelwidth}{1em}
        \setlength{\labelsep}{0.5em}
    }
}
\newcommand{\squishend}
{
    \end{list}
}

\title{Exploiting answer-invariant redundancies in satellite imagery for efficient VLM inference on edge}

\author{
  Ishani Janveja \quad Davis Zhang \quad Seoyul Oh \quad Deepak Vasisht \\
  University of Illinois Urbana-Champaign 
}

\begin{document}
\maketitle
\vspace{-0.3in}
\begin{abstract}
Onboard vision-language models could enable satellites to answer queries directly, but exhaustive tiled inference over high-resolution imagery is slow and energy-intensive. We identify answer-invariant token redundancy (AITR): image tiles and vision tokens that can be removed without changing the final answer. We present \name, a two-stage system that performs query-conditioned tile pruning followed by elastic prefill to reduce token budget. We evaluate it on LLaVA-1.5 7B running on Jetson AGX Orin. Compared with exhaustive tiled inference, \name\ reduces energy by 78\% and latency by 69\%, while increasing accuracy from 45\% to 73\%. 
\end{abstract}

\input{sections/intro-v2}
\input{sections/system_design_v2}
\input{sections/evaluation}
\input{sections/relatedwork}
\bibliographystyle{plain}
\bibliography{reference}
\appendix
\input{sections/appendix}

\end{document}

%% file: sections/intro-v2.tex
\section{Introduction}


The Earth has never been more richly probed. Today, large constellations of Low Earth Orbit (LEO) satellites can image the entire Earth multiple times every day. These satellites are transforming from just passive cameras in orbit around the Earth to active compute nodes equipped with edge-AI devices such as NVIDIA Jetson Orin or Thor~\cite{marshall2025owl, nvidia2026spacecomputing, singh2025starcloud, sriram2026spacex}. These devices are capable of running multimodal models with billions of parameters. Onboard AI enables a new vision where satellites can answer queries directly, without first downlinking imagery to ground servers. An analyst could ask, “How many boats are docked at this port right now?” Likewise, a wildfire-monitoring AI agent could query a constellation through an API and receive a report of active fires in California within minutes. 

However, such on-board query answering with vision-language models (VLMs) is infeasible today due to the scale of satellite imagery. Even a single query is far more expensive and expansive than a VLM's normal operating point. Satellite images covering just 2$\times$2~km$^2$ at 30~cm resolution spans roughly 6700$\times$6700 pixels. Models can only ingest images of smaller sizes. For instance, LLaVA1.5-7B~\cite{llavav1.5} which we run on Jetson Orin supports 336$\times$336 input image size. While downsizing is one option to process high-resolution satellite imagery, at that resolution, marine vessels become a handful of pixels, vehicles disappear and narrow flooded streets become visually ambiguous. Standard practice is therefore to tile the scene and process each independently. 

A 6700$\times$6700 pixel image is approximately 400 tiles. During inference, every tile is converted into a set of vision tokens (576 tokens, in case of LLaVA-1.5) which are processed by the language-model decoder as part of its input prompt. Thus, answering a query over all 400 tiles requires processing approximately 230K vision tokens in aggregate ($400\times576$). On a 64~GB Jetson AGX Orin, exhaustive inference takes 5--7 minutes per query. Latency over larger areas grows linearly to tens of minutes per image, while modern satellites like Planet Dove~\cite{planet2026planetscope} take $\approx$10$^4$ images per day~\cite{harwood2017minotaur}.  

In this paper, our primary goal is to decrease the energy and time spent per query by eliminating the fixed per-tile computation imposed by conventional VLM inference. We build \name\ (\textbf{R}edundancy s\textbf{ift}er) to make VLM inference efficient for satellite imagery by attacking  redundancy at two levels of granularity. Our key insight is that in satellite images, only a fraction of the tiles and vision tokens carry signal that support the answer to a query. Building on this insight, \name\ first uses \textit{tile pruning} to discard tiles that are unlikely to contain evidence relevant to the answer. At the token level, \name\ introduces \textit{elastic prefill}, a mechanism that dynamically allocates a vision-token budget to each retained tile. Both decisions are conditioned on the query and taken before the language model generates its first output token, preventing irrelevant visual content from consuming computation. \name\ makes both decisions to maximize the probability of a correct answer while reducing energy.

We evaluate \name\ on queries grounded in images from 2 large size satellite imagery datasets: xView and GLH-Bridge dataset. The energy and latency is benchmarked on a Jetson AGX Orin operating at the 60~W power mode to reflect the budget available on state-of-the-art satellites that carry edge-AI accelerators. Our evaluations on LLaVA-1.5-7B show that \name\ allows VLM inference to run on satellite edge at 78\% less energy and improves accuracy by 28 percent points by suppressing false positives compared to a system that does not reduce redundancy at all.

%% file: sections/system_design_v2.tex
\section{\name}

\subsection{Short Primer on VLM Inference}

VLM inference has four stages: vision encoding, modality projection, prefill, and decoding. Encoding converts an image into vision tokens and modality projection maps them into the language model's embedding space alongside the text tokens. During prefill, the model processes this entire input sequence and builds the KV cache before producing its first output token. Decoding then generates the remaining answer tokens one at a time. Intuitively, prefill is the cost of reading the input, while decoding is the cost of writing the answer.

\subsection{Answer-invariant Redundancy}

\begin{wrapfigure}{r}{0.6\textwidth}
    \centering
    \includegraphics[width=\linewidth]{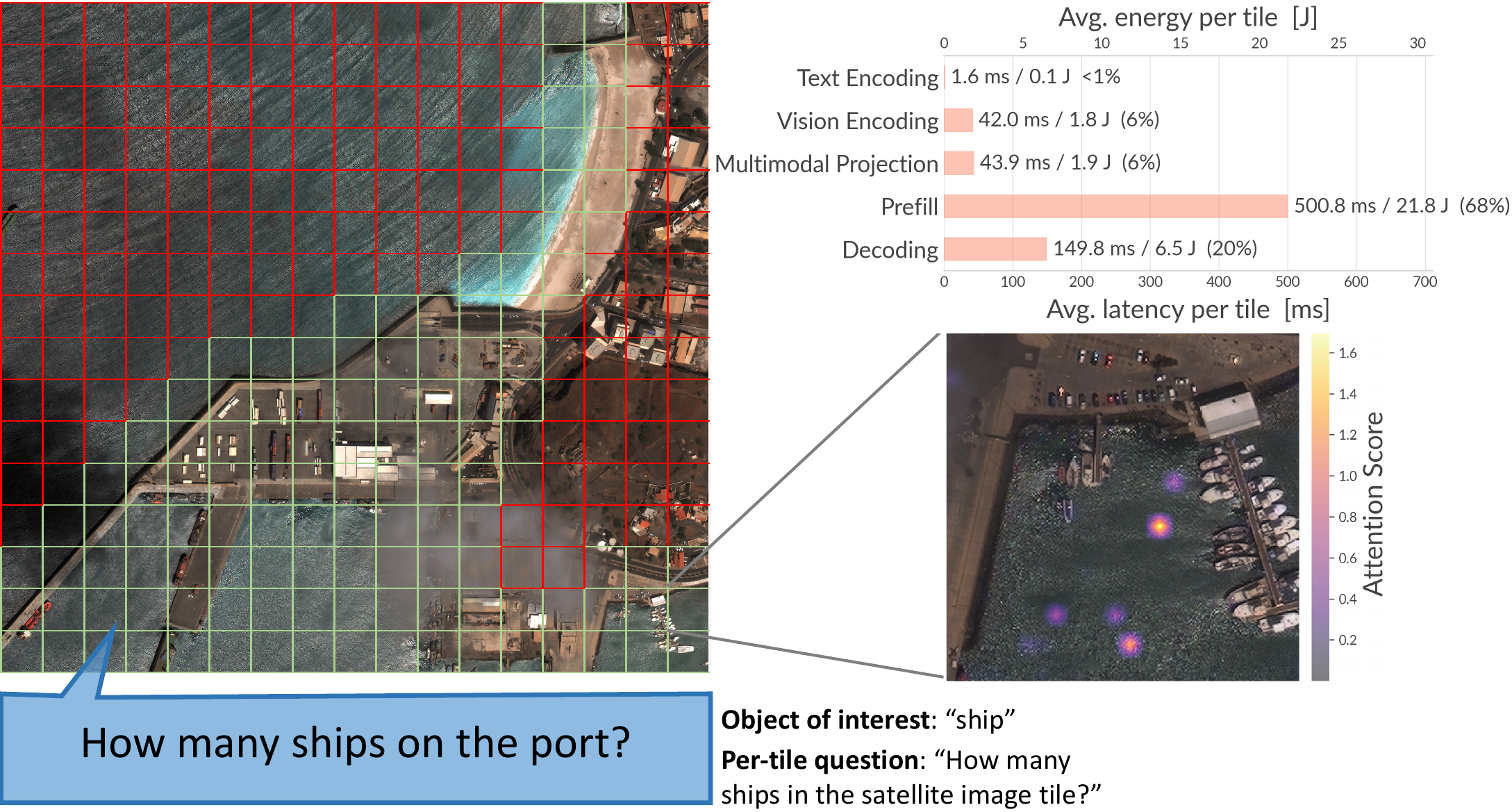}
    \caption{\textbf{\textit{Left:}} A tiled satellite scene and query; red boxes mark tiles irrelevant to the query. \textbf{\textit{Right (bottom):}} Attention is concentrated on a small subset of the vision tokens within one tile. \textbf{\textit{Right (top):}} Stage-level latency and energy, showing prefill dominates this input-heavy VLM inference. }
    \label{fig:aitr_prefilldominates}
    \vspace{-0.1in}
\end{wrapfigure}

Conventional tiled inference partitions a high-resolution satellite image into tiles and processes every query--tile pair with the VLM. The image encoder in the VLM emits a fixed number of vision tokens for each tile. In Fig.~\ref{fig:aitr_prefilldominates}~(left), for example, approximately 270 tiles produce 270 $\times$ 576 $\approx$156K vision tokens in aggregate on LLaVA-1.5. Across these invocations, the language model must prefill all of these tokens. However, much of this visual input is redundant. In Fig.\ref{fig:aitr_prefilldominates}~(left), many tiles contain only ocean or urban background. Even within a tile, attention is concentrated on only a small subset of its vision tokens, as shown in Fig.\ref{fig:aitr_prefilldominates}~(right, bottom). We call this \emph{answer-invariant tile and token redundancy (AITR)}: visual content that can be removed, at either tile or token granularity, without changing the answer.

AITR makes prefill costly in satellite images. While LLM queries are typically decode-dominated (a short prompt, then a long generation), satellite VLM queries reverse the trend. They have enormous visual inputs but answers of only a few words, therefore inference is prefill-dominated. Here vision-token count determines both latency and energy per query. Fig.\ref{fig:aitr_prefilldominates}~(right top) shows the split in latency and energy for different VLM stages when prompting a satellite imagery.

\subsection{System Design}
\label{sec:design}
\begin{figure}
    \centering
    \includegraphics[width=\linewidth]{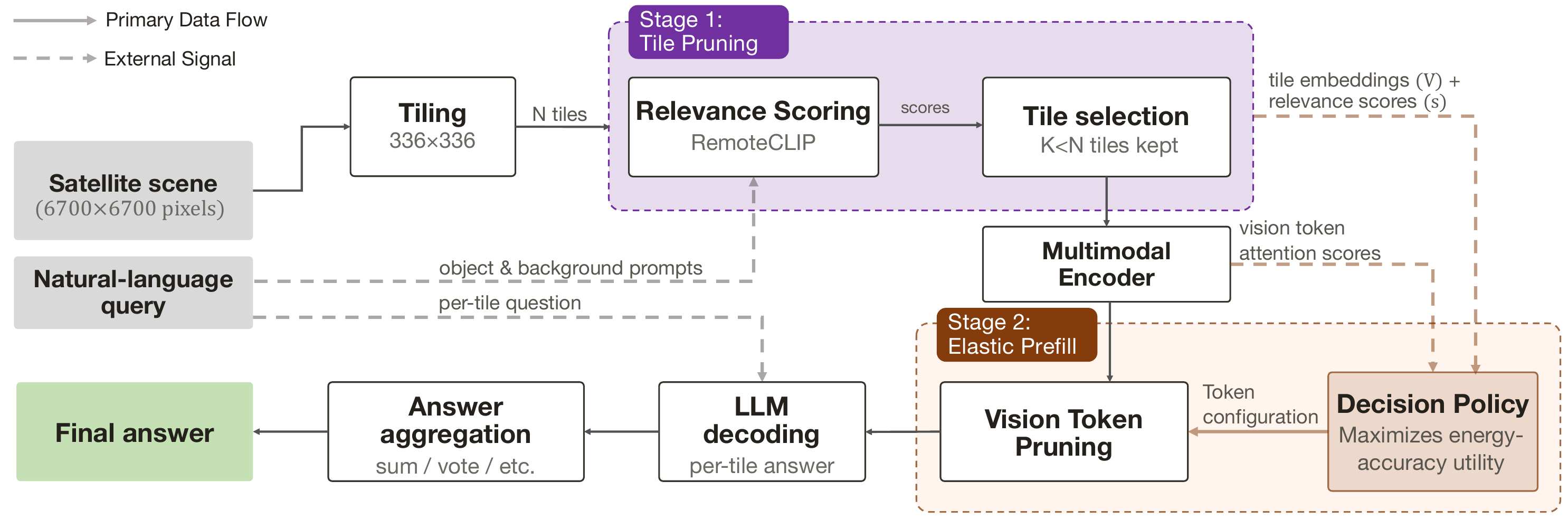}
    \vspace{-0.2in}
    \caption{\name\ reduces answer-invariant tile and token redundancy with its 2-stage framework:\\ (i) Tile Pruning, and (ii) Elastic Prefill.}
    \vspace{-0.1in}
    \label{fig:e2e_pipeline}
\end{figure}

\name's framework, shown in Fig.\ref{fig:e2e_pipeline}, uses a two-fold approach to address AITR. Appendix~\ref{app:walkthrough} walks through the full pipeline on a real test query.


\textbf{Stage 1 -- Tile Pruning:} \name\ starts by assigning a query-conditioned relevance score to every tile and pruning out ones that do not survive a set threshold $\tau$. Consider the query in Fig.\ref{fig:aitr_prefilldominates} where the object of interest in the shown query is ``ship''. \name\ uses a CLIP-style multimodal encoder to embed each tile, the query’s object of interest (``ship''), and a fixed vocabulary of common satellite background terms (``open water'', ``bare ground'', ``clouds'', \ldots) into a shared space. Let $V_i$, $t$ and $\{B_j\}_{j=1}^{M}$ denote their $\ell_2$-normalized embeddings. We define the relevance score of tile $i$ as the margin between its cosine similarity to the object of interest and the background vocabulary:
\begin{equation}
s_i = V_i t - \frac{1}{M} \sum_{j=1}^{M} V_i B_j^{\top},
\qquad i \in [N].
\label{eq:relevance_contrast_margin}
\end{equation}
Subtracting background similarity reduces context-driven false positives. For example, an open-water tile may score highly for ``ship'' because CLIP associates ships with water. The contrastive score instead favors tiles that match the target more strongly than common backgrounds. Finally, \name{} min--max normalizes scores within each
scene and retains tiles whose normalized score exceeds $\tau$.




\textbf{Stage 2 -- Elastic Prefill:} For each retained tile from stage 1, the vision encoder of the VLM emits a fixed number of vision tokens $n_v$, set by the encoder configuration. Because attention is concentrated on a small subset of vision tokens, we use VisionZip's compression mechanism~\cite{visionzip} to retain dominant high-attention tokens and merge the remainder. However, the key challenge is that different query--tile pairs contain different amounts of answer-invariant redundancy and therefore tolerate different compression levels. We introduce elastic prefill to solve this by choosing $k$ independently for each tile to balance expected answer accuracy against energy cost.

In practice, \name\ selects each tile's token budget from a discrete set $\mathcal{C}$. For LLaVA-1.5, whose vision encoder emits 576 tokens per tile, we use $\mathcal{C}=\{576,432,288,144,72\}$. For tile $i$, the decision policy assigns each candidate budget $c\in\mathcal{C}$ the utility:
\begin{equation}
U_i(c)=\widehat{A}_i(c)-\lambda E_c,
\label{eq:elastic-utility}
\end{equation}
where $\widehat{A}_i(c)$ is the expected answer accuracy under budget $c$, $E_c$ is its average energy cost, and $\lambda$ controls the accuracy--energy trade-off. We profile $E_c$ offline using \texttt{tegrastats}~\cite{nvidia_tegrastats} for each value of $c$. The central challenge is estimating the tile-specific accuracy under each budget before the VLM generates an answer.


The key observation is that a tile that does not contain the target object can usually tolerate aggressive compression (high expected accuracy at small $c$). Whereas, a tile containing the target may need
more tokens depending on task type (e.g., counting vs. identification), the object’s visibility, and its surrounding context. \name\ captures this offline: for each token configuration $c\in \mathcal{C}$, it profiles the VLM’s accuracy separately on object-containing tiles ($\alpha^{+}_c$) and object-absent tiles ($\alpha^{-}_c$)

At runtime, \name\ does not know whether a retained tile contains the target. It therefore estimates this. We train a gradient-boosted ensemble of decision trees to predict $p_i$, the probability that tile $i$ contains the object of interest. The predictor uses features already available from the pipeline, including the tile embedding $V_i$, its Stage~1 relevance score, vision-token attention scores, and the relevance scores of its four quadrants. The expected accuracy of processing tile $i$ under configuration $c$ is then approximated as: $\widehat{A}_i(c) = p_i\,\alpha^{+}_c + (1 - p_i)\,\alpha^{-}_c$. 

Finally, elastic prefilling selects, for each tile, the configuration that best trades expected accuracy against energy:$c^\star_i = \operatorname*{arg\,max}_{c \in \mathcal{C}}\; U_i(c)$
Each retained tile is processed at its selected configuration, and the per-tile answers are aggregated into the final response: counting queries sum the per-tile counts, while identification queries are voted ``yes'' if $\geq$2 tiles return positive answers.



%% file: sections/evaluation.tex
\section{Evaluation}
\textbf{\textit{Setup.}} We benchmark \name\ on Jetson AGX Orin running at 60W power mode. Our dataset consists of visual question answering (VQA) queries on two high-resolution, large satellite image datasets: xView (with image size 2576$\times$2426 to 5121$\times$3325 pixels) and GLH-Bridge dataset (2048$\times$2048 to 8192$\times$8192 pixels). These datasets consist of object labels and annotations in satellite images. Therefore, we adapt them for VQA by generating whole-image existence and counting questions over the images. Refer to appendix~\ref{app:dataset} for details. Finally, to implement \name\ stage-1 (tile pruning) we use RemoteCLIP~\cite{liu2024remoteclip}, a vision-language model trained on satellite imagery. Stage-2 relies on LLaVA-1.5 (7B) for processing the VQA on the retained tiles. 
\begin{wrapfigure}{r}{0.7\textwidth}
    \centering
    \includegraphics[width=1\linewidth]{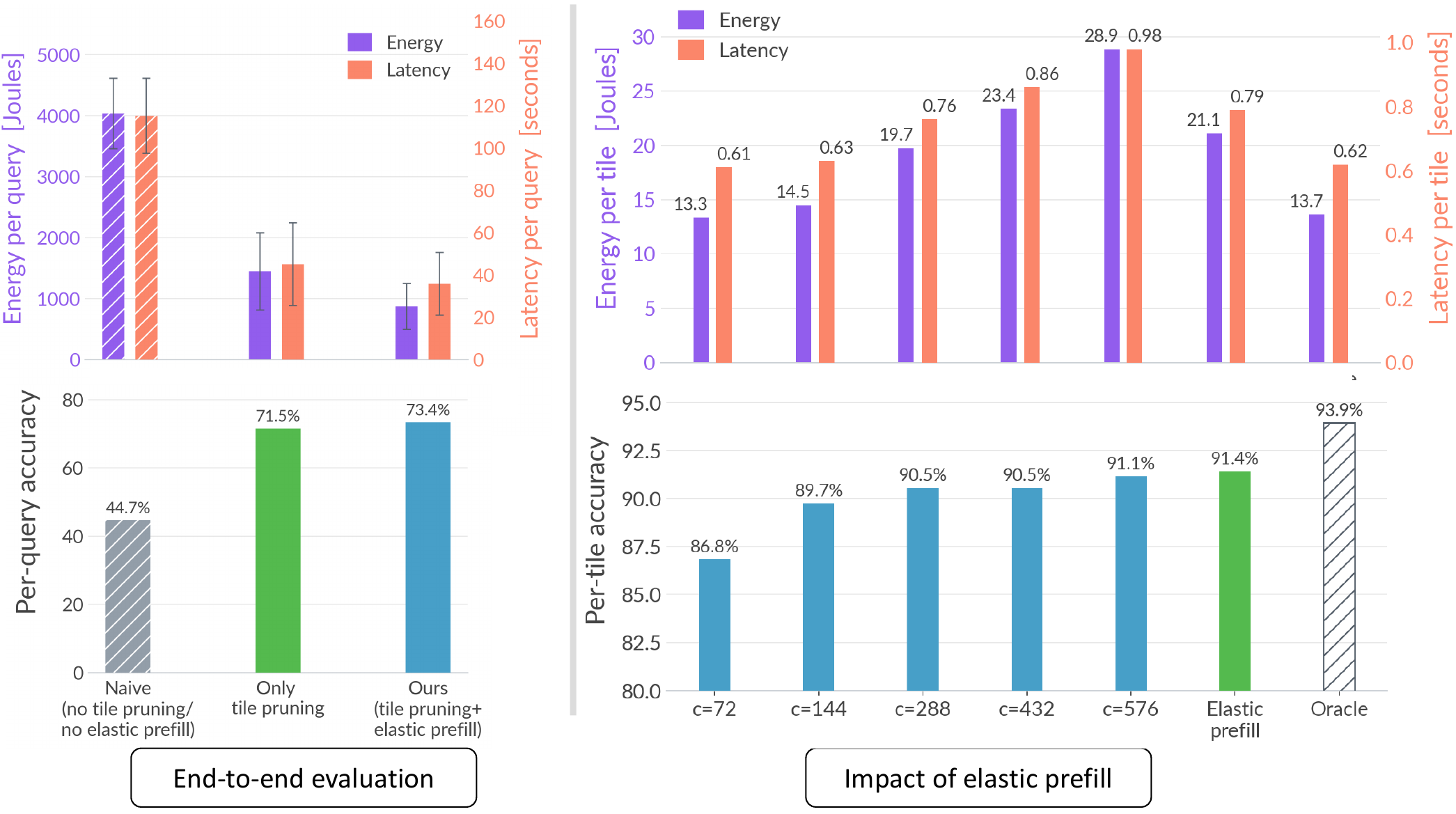}
    \caption{(\textbf{\textit{Left}}) Avg. per-query energy, latency and accuracy of \name\ compared with a naive system and one that only prunes out redundant tiles. (\textbf{\textit{Right}}) Impact of elastic prefill: Comparing avg. vlm inference energy, latency and accuracy with fixed configurations per-tile.}
    \label{fig:eval}
    \vspace{-0.2in}
\end{wrapfigure}

\textbf{\textit{Benchmarks.}} Fig.\ref{fig:eval}(left) compares \name\ with existing approaches do not deal with redundancy. Our system saves {78\%} energy and {69\%} latency compared to such a naive deployment that does not eliminate AITR. Tile pruning alone contributes to {64\%} energy and {60\%} latency savings and elastic prefill adds additional 14\% energy and 9\% latency savings. To generate these benchmarks we set the normalized relevance score threshold ($\tau=0.37$). The threshold was deliberately kept low to favor recall (81\%) over precision (17\%) since accidentally discarded tiles cannot be seen by the VLM. A falsely retained tile costs only additional inference energy, and elastic prefill bounds that cost by assigning low-relevance tiles few vision tokens. Accuracy also improves with redundancy removal. Tile pruning particularly reduces false positives by filtering out negative tiles from reaching the VLM.

Next, we evaluate the efficacy of \name's elastic prefill. We try to answer the question that if we were to simply discard a fixed number of vision tokens during inference (similar to VisionZip~\cite{visionzip}) would that be enough? From Fig.\ref{fig:eval}(right) it is evident that elastic prefill the cheapest and most accurate compared to fixed configurations. We also benchmark elastic prefill against an ``oracle'' policy that, given the ground-truth answer, picks the cheapest configuration per tile that still answers correctly. The oracle saves {33\%} more energy compared to our current elastic prefill decision policy and increases accuracy by 3\%. This sets a ceiling for how much a better per-tile predictor could buy.


%% file: sections/relatedwork.tex
\section{Related Work}



\textbf{\textit{A. Satellite edge query processing.}} Prior systems reduce satellite query latency and downlinked bytes through adaptive onboard processing~\cite{kodan, serval, earthsight}. 
These systems optimize query placement, scheduling, and data transmission. \name\ is complementary in that once a VLM query is assigned to a satellite, it reduces its inference energy by pruning irrelevant image tiles and allocating vision--token budgets across the retained tiles. \textbf{\textit{B. Efficient VLM inference.}} A growing body of work reduces VLM inference cost by exploiting redundancy in vision tokens. While some of them target compression before prefill~\cite{visionzip}, others target pruning in the LLM layers during prefill~\cite{chen2024image,zhang2024sparsevlm}. There are also systems that adapt to input rather than imposing a single fixed compression ratio~\cite{yeatpllava2025,li2026catp,11445457}. Most closely related to our setting is~\cite{luo2025iccv}, that combines text-guided tile selection with vision-token pruning for large remote-sensing images. However, prior methods optimize computational proxies such as token count, FLOPs, or latency rather than total consumed energy on a edge platform. \name\ profiles the energy and accuracy of candidate token budgets and uses these measurements to allocate vision-token budgets to each query--tile pair.

%% file: sections/appendix.tex
\section{Dataset Generation}
\label{app:dataset}
We constructed two question-answering (QA) datasets from existing remote-sensing object annotations: xView~\cite{lam2018xview} and GLH-Bridge~\cite{li2024glhbridge}. In both datasets, questions are generated from fixed templates and ground-truth answers are derived deterministically from the source annotations. We consider two tasks: (1) object identification and (2) object counting. 

\paragraph{Object identification.}
The identification task asks whether at least one instance of a target semantic category is present in the image, with a binary answer space of \textit{Yes} or \textit{No}. For example:
\begin{quote}
\textit{Are any instances of aircraft visible in this satellite image? Answer Yes or No.}
\end{quote}

\paragraph{Object counting.}
The object-counting task asks which of four ranges contains the total number of target instances in the image: \textbf{0}, \textbf{1--2}, \textbf{3--9}, and \textbf{more than 9}. For example:
\begin{quote}
\textit{How many instances of aircraft are visible in this satellite image? Choose one: 0, 1--2, 3--9, or more than 9.}
\end{quote}

We retain zero-count examples as part of the counting task. Although these examples necessarily indicate the absence of the target category, they constitute a distinct count bucket and allow us to evaluate whether models spuriously predict objects in target-absent images.

\subsection{xView}

\paragraph{Source data and semantic categories.}
We use publicly annotated xView training images and their corresponding GeoJSON annotations. The original 60 xView annotation classes contain distinctions that are unnecessarily fine-grained for our evaluation. We therefore group related classes into broader semantic categories that can be expressed naturally in whole-image visual questions. For example, \textit{fixed-wing aircraft}, \textit{small aircraft}, \textit{passenger/cargo plane}, and \textit{helicopter} are grouped as \textit{aircraft}. 

We use four categories with sufficient positive and negative annotation support for balanced object identification: aircraft, boats or ships, buildings or structures, and rail vehicles. For object counting, we use aircraft, boats or ships, and rail vehicles, excluding buildings or structures because dense and partially connected structures can make instance boundaries visually ambiguous.

For each image $i$ and semantic category $c$, we compute the whole-image object count $N(i,c)$ by summing all xView annotations mapped to $c$, and derive the corresponding identification and counting labels using the definitions above.

\paragraph{Identification questions.}
For each category, we construct positive examples from image--category pairs with $N(i,c)>0$ and negative examples from pairs with $N(i,c)=0$. To avoid trivial negatives from otherwise empty scenes, negative examples are restricted to images containing at least one annotated object from another xView category. We balance positive and negative examples independently within each category by subsampling the majority class. This yields 978 identification questions: 276 for aircraft, 318 for boats or ships, 232 for buildings or structures, and 152 for rail vehicles, spanning 635 unique images.

\paragraph{Counting questions.}
For each counting category, we balance the four count buckets by subsampling each bucket to the size of the smallest bucket. Zero-count examples follow the same negative-example criterion as the identification task. This yields 172 aircraft, 152 boats-or-ships, and 48 rail-vehicle questions, for 372 counting questions in total.

Overall, the xView dataset contains 1,350 whole-image questions over 656 unique images.

\subsection{GLH-Bridge}
\paragraph{Source data.}
We use the publicly annotated GLH-Bridge training and validation images and their corresponding bridge annotations. Unlike xView, GLH-Bridge contains a single target category, \textit{bridge}, so no semantic class aggregation is required. For each image $i$, we compute the whole-image bridge count $N(i)$ as the number of annotated bridge instances and derive the corresponding identification and counting labels using the definitions above.

\paragraph{Identification questions.}
We construct positive examples from images with $N(i)>0$ and negative examples from images with $N(i)=0$. Because the number of zero-bridge images is substantially smaller than the positive pool, we retain all 99 negative images and subsample 99 positive images. This yields 198 balanced identification questions.

\paragraph{Counting questions.}
We group images into the four count buckets defined above and balance the buckets by subsampling each to the size of the smallest bucket. The zero-count bucket is limiting, with 99 images, so this yields 396 counting questions, with 99 examples per bucket.

Overall, the GLH-Bridge dataset contains 594 whole-image questions over 489 unique images.

\section{End-to-end walkthrough on a test query from xView image 1565.tif}
Here we trace one test query through the deployed pipeline. All scores, budgets, and answers below are taken from an actual run of the system.
\label{app:walkthrough}
\textbf{User Query:} 
``How many instances of aircraft are visible? Choose one: 0, 1--2, 3--9, or more than 9'' <1565.tif>

\textbf{Ground Truth:} 3--9

\begin{figure}[!htbp]
    \raggedright
    \includegraphics[width=0.4\linewidth]{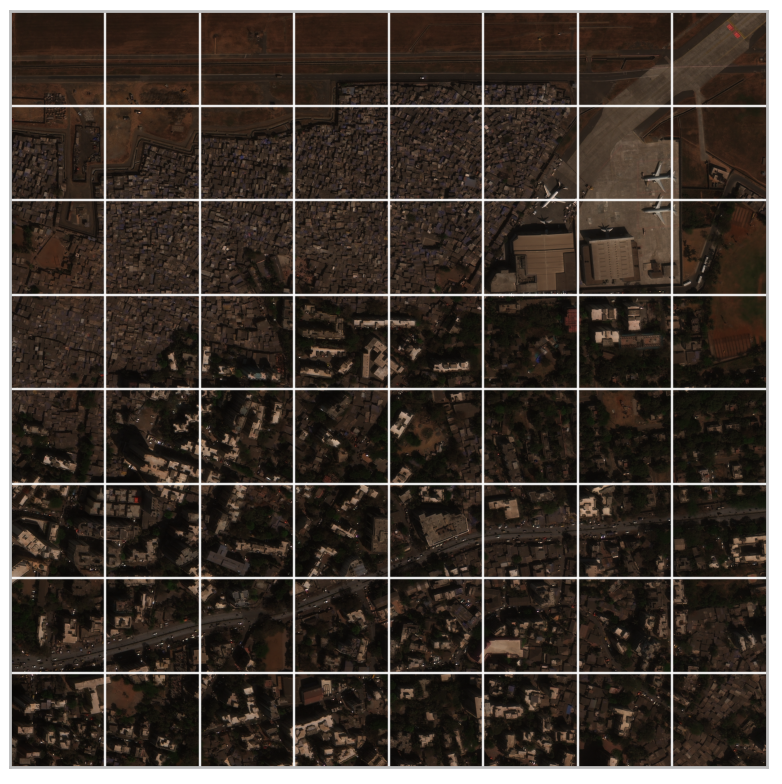}
    \captionsetup{justification=raggedright,singlelinecheck=false}
    \caption{Step~1. Tile the region $8{\times}8$ tiles of $336^2$ px}
    \vspace{-0.1in}
    \label{fig:e2ewalk_1}
\end{figure}

\textbf{Step 1: Tiling.}~The $2960{\times}2814$\,px region is cropped into an $8{\times}8$ grid of $336^2$\,px tiles, matching the VLM's native input resolution, giving $N=64$ tiles.(Fig.\ref{fig:e2ewalk_1})


\begin{figure}[!htbp]
\raggedright
    \includegraphics[width=0.4\linewidth]{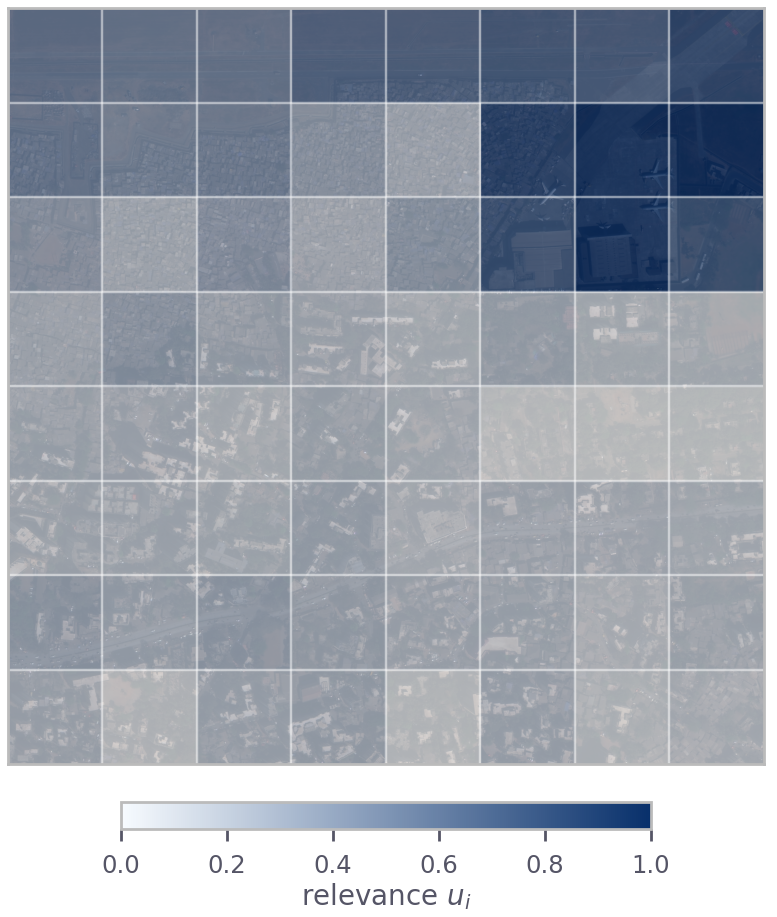}
    \captionsetup{justification=raggedright,singlelinecheck=false}
    \caption{Step~2. Per-tile relevance score}
    \label{fig:e2ewalk_2}
\end{figure}

\textbf{Step 2: Relevance scoring.} Stage~1 encodes every tile once with
RemoteCLIP and scores it against the query object (``aircraft'') using a
background-margin score: the cosine similarity to the object phrase minus the mean similarity to a fixed set of generic background terms (Listing~\ref{lst:relevance}). Each tile's score is then smoothed against its $m{=}6$ most visually similar tiles and range-normalized within the image to $u_i \in [0,1]$. The aircraft cluster in the top-right corner scores highest; the dense urban fabric in the lower two-thirds scores near the bottom of the range.

\begin{figure}[!htbp]
    \raggedright
    \includegraphics[width=0.4\linewidth]{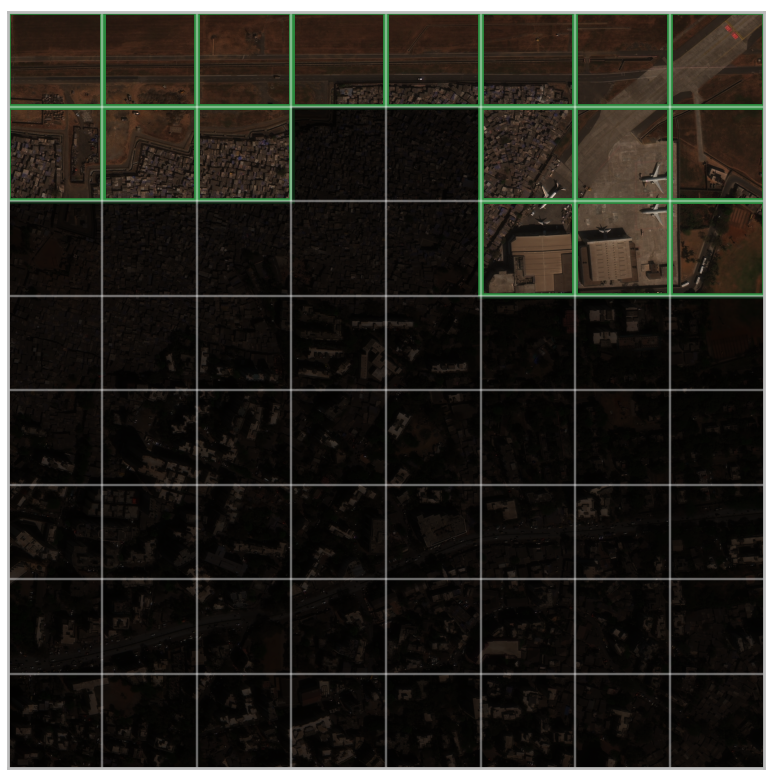}
    \captionsetup{justification=raggedright,singlelinecheck=false}
    \caption{Step 3. Prune irrelevant tiles}
    \vspace{-0.1in}
    \label{fig:e2ewalk_3}
\end{figure}

\para{Step 3: Tile pruning.} Tiles with $u_i \geq 0.37$ are retained---17 of 64 here, a $73\%$ reduction in VLM calls before any token pruning occurs. Because scores are normalized per-tile and the top-scoring tile always has $u_i = 1$, no query can be left with zero tiles.

\begin{figure}[!htbp]
    \raggedright
    \includegraphics[width=0.4\linewidth]{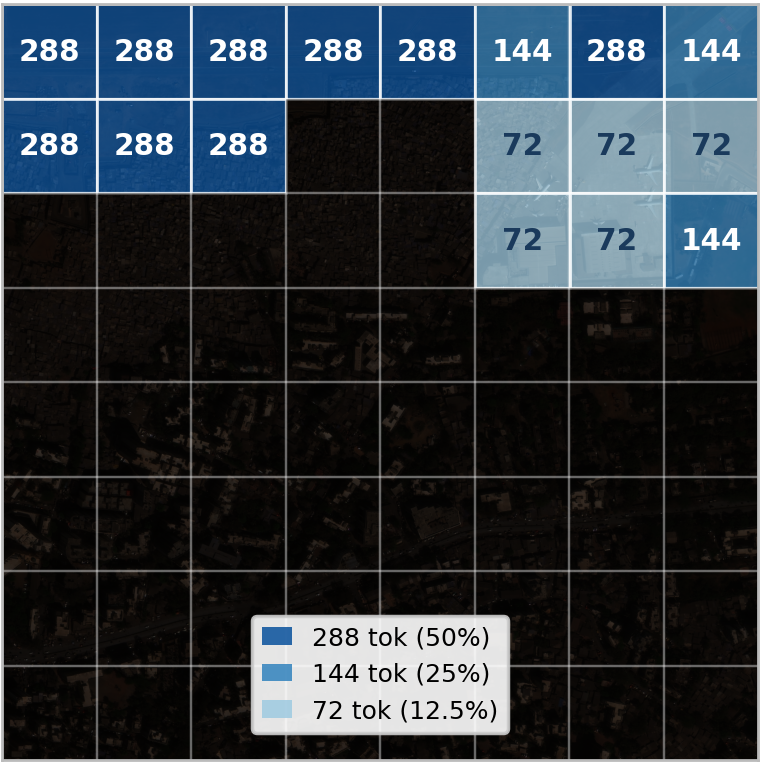}
    \captionsetup{justification=raggedright,singlelinecheck=false}
    \caption{Step 4. Elastic prefill decision policy picks per tile token budget}
    \vspace{-0.1in}
    \label{fig:e2ewalk_4}
\end{figure}


\textbf{Step 4: Per-tile token budgets.} For each retained tile, \name's utility-maximizing decision policy evaluates the utility of Eq.~\eqref{eq:elastic-utility} and selects $c^\star_i \in \mathcal{C}$. The resulting assignment may look inverted at first glance: the tiles that actually contain aircraft get the \emph{smallest} budget (72 tokens), while empty urban tiles get 288. This is Eq.~\eqref{eq:elastic-utility} working as intended. For the aircraft tiles the predictor is confident ($p_i$ high), so $\widehat{A}_i(c) \approx \alpha^+_c$---and the offline profiles show that compressing a tile with a clearly visible target object barely changes the count the VLM reports, so extra tokens buy no accuracy and the policy spends the minimum. For the clutter tiles $p_i$ is low, so $\widehat{A}_i(c) \approx \alpha^-_c$---which \emph{does} degrade under aggressive compression: over-compressed empty tiles hallucinate objects, and on a counting query every false count is added directly into the final sum. The policy therefore pays for tokens exactly where compression would corrupt the answer, not where the objects are.

\begin{figure}[!htbp]
    \raggedright
    \includegraphics[width=0.4\linewidth]{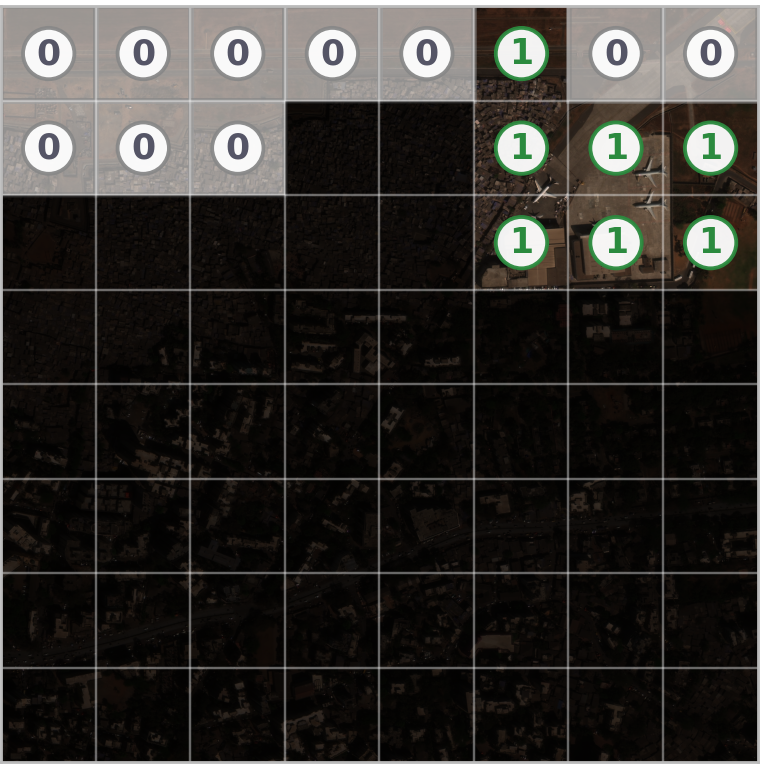}
    \captionsetup{justification=raggedright,singlelinecheck=false}
    \caption{Step 5. Per-tile VLM answers $\Sigma$ counts $= 7 \rightarrow$ ``3–9'' (GT: 3–9)}
    \vspace{-0.1in}
    \label{fig:e2ewalk_5}
\end{figure}

\textbf{Step 5: Per-tile answers and aggregation.} Each retained tile is answered independently (``How many aircraft are in this tile? Answer with a single integer''). Seven tiles report a count of 1; the counts are summed ($\Sigma = 7$) and mapped to the bucket ``3--9'', matching the ground truth. In total, the query used 17 VLM calls at an average of 36\% of the full token budget, and no call at all on the other 47 tiles.

\begin{lstlisting}[language=Python, caption={Stage-1 relevance scoring \texttt{E(...)} is the RemoteCLIP encoder returning unit-norm embeddings; notation follows \S~\ref{sec:design}}, label={lst:relevance}]
  V = E(tiles)  # (N, d) tile embeddings
  t = E("a satellite image of aircraft")    # object of interest
  B = E(["open water", "bare ground", "clouds", ...])   # (M, d) background
  
  s = V @ t - (V @ B.T).mean(axis=1)  # Eq. (1): background margin
  # impl. detail: t averages 5 caption templates; each s_i is then
  # smoothed with its 6 most similar tiles to suppress per-tile noise
  u = (s - s.min()) / (s.max() - s.min())   # min-max normalize per scene
  retained = [i for i in range(N) if u[i] >= 0.37]   # threshold tau
\end{lstlisting}

%% file: reference.bib
@misc{marshall2025owl,
  author       = {Will Marshall},
  title        = {Introducing Owl: Planet's Most Advanced Satellite Mission Yet},
  year         = {2025},
  month        = oct,
  day          = {7},
  howpublished = {Planet Pulse},
  url          = {https://www.planet.com/pulse/introducing-owl-planet-s-most-advanced-satellite-mission-yet/},
  note         = {Accessed: 2026-07-24}
}

@misc{nvidia2026spacecomputing,
  author       = {{NVIDIA Corporation}},
  title        = {{NVIDIA} Launches Space Computing, Rocketing {AI} Into Orbit},
  year         = {2026},
  month        = mar,
  day          = {16},
  howpublished = {NVIDIA Newsroom},
  url          = {https://nvidianews.nvidia.com/news/space-computing},
  note         = {Accessed: 2026-07-24}
}

@misc{sriram2026spacex,
  author       = {Akash Sriram},
  title        = {{SpaceX} Aims to Launch Orbital {AI} Computing Tests by End of Next Year, Sources Say},
  year         = {2026},
  month        = jun,
  day          = {9},
  howpublished = {Reuters},
  url          = {https://www.reuters.com/business/media-telecom/spacex-aims-launch-orbital-ai-computing-tests-by-end-next-year-sources-say-2026-06-09/},
  note         = {Accessed: 2026-07-24}
}

@misc{singh2025starcloud,
  author       = {Pia Singh},
  title        = {`Greetings, Earthlings': {Nvidia}-Backed {Starcloud} Trains First {AI} Model in Space as Orbital Data Center Race Heats Up},
  year         = {2025},
  month        = dec,
  day          = {10},
  howpublished = {CNBC},
  url          = {https://www.cnbc.com/2025/12/10/nvidia-backed-starcloud-trains-first-ai-model-in-space-orbital-data-centers.html},
  note         = {Accessed: 2026-07-24}
}

@article{earthsight,
  title={EarthSight: A Distributed Framework for Low-Latency Satellite Intelligence},
  author={Erol, Ansel Kaplan and Lee, Seungjun and Mahajan, Divya},
  journal={arXiv preprint arXiv:2511.10834},
  year={2025}
}

@inproceedings{kodan,
  title={Kodan: Addressing the computational bottleneck in space},
  author={Denby, Bradley and Chintalapudi, Krishna and Chandra, Ranveer and Lucia, Brandon and Noghabi, Shadi},
  booktitle={Proceedings of the 28th ACM International Conference on Architectural Support for Programming Languages and Operating Systems, Volume 3},
  pages={392--403},
  year={2023}
}

@inproceedings{serval,
  title={Known knowns and unknowns: Near-realtime earth observation via query bifurcation in serval},
  author={Tao, Bill and Chabra, Om and Janveja, Ishani and Gupta, Indranil and Vasisht, Deepak},
  booktitle={21st USENIX Symposium on Networked Systems Design and Implementation (NSDI 24)},
  pages={809--824},
  year={2024}
}

@inproceedings{visionzip,
    author    = {Yang, Senqiao and Chen, Yukang and Tian, Zhuotao and Wang, Chengyao and Li, Jingyao and Yu, Bei and Jia, Jiaya},
    title     = {VisionZip: Longer is Better but Not Necessary in Vision Language Models},
    booktitle = {Proceedings of the IEEE/CVF Conference on Computer Vision and Pattern Recognition (CVPR)},
    month     = {June},
    year      = {2025},
    pages     = {19792-19802}
}

@article{liu2024remoteclip,
  title={Remoteclip: A vision language foundation model for remote sensing},
  author={Liu, Fan and Chen, Delong and Guan, Zhangqingyun and Zhou, Xiaocong and Zhu, Jiale and Ye, Qiaolin and Fu, Liyong and Zhou, Jun},
  journal={IEEE Transactions on Geoscience and Remote Sensing},
  volume={62},
  pages={1--16},
  year={2024},
  publisher={IEEE}
}

@misc{llavav1.5,
      title={Improved Baselines with Visual Instruction Tuning}, 
      author={Haotian Liu and Chunyuan Li and Yuheng Li and Yong Jae Lee},
      year={2024},
      eprint={2310.03744},
      archivePrefix={arXiv},
      primaryClass={cs.CV},
      url={https://arxiv.org/abs/2310.03744}, 
}

@InProceedings{luo2025iccv,
    title={When Large Vision-Language Model Meets Large Remote Sensing Imagery: Coarse-to-Fine Text-Guided Token Pruning},
    author={Luo, Junwei and Zhang, Yingying and Yang, Xue and Wu, Kang and Zhu, Qi and Liang, Lei and Chen, Jingdong and Li, Yansheng},
    booktitle={Proceedings of the IEEE/CVF International Conference on Computer Vision (ICCV)},
    month={October},
    year={2025},
    pages={9206-9217}
}

@misc{chen2024image,
      title={An Image is Worth 1/2 Tokens After Layer 2: Plug-and-Play Inference Acceleration for Large Vision-Language Models}, 
      author={Liang Chen and Haozhe Zhao and Tianyu Liu and Shuai Bai and Junyang Lin and Chang Zhou and Baobao Chang},
      year={2024},
      eprint={2403.06764},
      archivePrefix={arXiv},
      primaryClass={cs.CV}
}

@inproceedings{zhang2024sparsevlm,
  title={SparseVLM: Visual Token Sparsification for Efficient Vision-Language Model Inference},
  author={Zhang, Yuan and Fan, Chun-Kai and Ma, Junpeng and Zheng, Wenzhao and Huang, Tao and Cheng, Kuan and Gudovskiy, Denis and Okuno, Tomoyuki and Nakata, Yohei and Keutzer, Kurt and others},
  booktitle={International Conference on Machine Learning},
  year={2025}
}

@InProceedings{yeatpllava2025,
    author    = {Ye, Xubing and Gan, Yukang and Ge, Yixiao and Zhang, Xiao-Ping and Tang, Yansong},
    title     = {ATP-LLaVA: Adaptive Token Pruning for Large Vision Language Models},
    booktitle = {Proceedings of the IEEE/CVF Conference on Computer Vision and Pattern Recognition (CVPR)},
    month     = {June},
    year      = {2025},
    pages     = {24972-24982}
}

@misc{nvidia_tegrastats,
  author       = {{NVIDIA Corporation}},
  title        = {{Tegrastats Utility}},
  howpublished = {\emph{NVIDIA Jetson Linux Developer Guide}, Release 36.4.4},
  year         = {2026},
  url          = {https://docs.nvidia.com/jetson/archives/r36.4.4/DeveloperGuide/AT/JetsonLinuxDevelopmentTools/TegrastatsUtility.html},
  note         = {Last updated January 16, 2026; accessed August 28, 2026}
}

@inproceedings{li2026catp,
  title={Catp: Contextually adaptive token pruning for efficient and enhanced multimodal in-context learning},
  author={Li, Yanshu and Yang, Jianjiang and Shen, Zhennan and Han, Ligong and Xu, Haoyan and Tang, Ruixiang},
  booktitle={Proceedings of the AAAI Conference on Artificial Intelligence},
  volume={40},
  number={8},
  pages={6619--6627},
  year={2026}
}

@INPROCEEDINGS{11445457,
  author={Yang, Sihan and Xu, Runsen and Cui, Chenhang and Wang, Tai and Lin, Dahua and Pang, Jiangmiao},
  booktitle={2025 IEEE/CVF International Conference on Computer Vision (ICCV)}, 
  title={VFLowOpt: A Token Pruning Framework for LMMs with Visual Information Flow-Guided Optimization}, 
  year={2025},
  volume={},
  number={},
  pages={23924-23934},
  doi={10.1109/ICCV51701.2025.02218}}

@online{harwood2017minotaur,
  author       = {Harwood, William},
  title        = {Minotaur Rocket Launches Earth-Observation Satellites},
  organization = {CBS News},
  date         = {2017-10-31},
  url          = {https://www.cbsnews.com/news/minotaur-rocket-launches-10-earth-observation-satellites-planet/},
  urldate      = {2026-08-28}
}

@misc{li2024glhbridge,
  author={Li, Yansheng and Luo, Junwei and Zhang, Yongjun and Tan, Yihua and Yu, Jin-Gang and Bai, Song},
  journal={IEEE Transactions on Pattern Analysis and Machine Intelligence}, 
  title={Learning to Holistically Detect Bridges From Large-Size VHR Remote Sensing Imagery}, 
  year={2024},
  volume={46},
  number={12},
  pages={11507-11523},
  doi={10.1109/TPAMI.2024.3393024}}

@misc{lam2018xview,
  title={xview: Objects in context in overhead imagery},
  author={Lam, Darius and Kuzma, Richard and McGee, Kevin and Dooley, Samuel and Laielli, Michael and Klaric, Matthew and Bulatov, Yaroslav and McCord, Brendan},
  journal={arXiv preprint arXiv:1802.07856},
  year={2018}
}

@misc{planet2026planetscope,
  author       = {{Planet Labs PBC}},
  title        = {PlanetScope},
  year         = {2026},
  howpublished = {\url{https://docs.planet.com/data/imagery/planetscope/\#psbsd}},
  note         = {Planet Documentation, PSB.SD section.
                  Last updated July 30, 2026;
                  accessed August 29, 2026}
}
